\pdfoutput=1

\documentclass[11pt]{article}

\usepackage[]{acl}

\usepackage{times}
\usepackage{latexsym}

\usepackage[T1]{fontenc}

\usepackage[utf8]{inputenc}

\usepackage{microtype}

\usepackage{inconsolata}

\usepackage{graphicx}

\usepackage{amsmath}
\usepackage{amsthm}
\usepackage{amsfonts}
\usepackage{booktabs}
\usepackage{multirow}
\usepackage{xspace}

\newcommand{\mistral}{\texttt{mistral-7b}\xspace}
\newcommand{\tulu}{\texttt{tulu-2-7b}\xspace}
\newcommand{\mistralref}{\texttt{mistral-v0.2}\xspace}
\newcommand{\llama}{\texttt{llama3}\xspace}

\title{Understanding Reference Policies in Direct Preference Optimization}

\author{
 Yixin Liu$^{1}$ 
  \quad \textbf{Pengfei Liu}$^{2}$ 
 \quad \textbf{Arman Cohan}$^{1,3}$ \vspace{6pt}\\
  $^1$Yale University\quad 
  $^2$Shanghai Jiao Tong University 
  \quad
  $^3$Allen Institute for AI
  \vspace{6pt}\\
  \texttt{yixin.liu@yale.edu, pengfei@sjtu.edu.cn, arman.cohan@yale.edu}
 }

\begin{document}
\maketitle
\begin{abstract}
Direct Preference Optimization (DPO) has become a widely used training method for the instruction fine-tuning of large language models (LLMs). 
In this work, we explore an under-investigated aspect of DPO -- its dependency on the reference model or policy.
Such reference policies, typically instantiated as the model to be further fine-tuned, are important since they can impose an upper limit on DPO's effectiveness.
Therefore, we address three related research questions in this work.
First, we explore the optimal strength of the KL divergence constraint in DPO, which penalizes deviations from the reference policy, and find that DPO is sensitive to this strength.
Next, we examine the necessity of the KL-constraint from the reference policies in DPO by providing both theoretical and empirical comparisons between DPO and related learning objectives, demonstrating DPO's superiority in this controlled setting.
Additionally, we investigate whether DPO benefits from stronger reference policies, finding that a stronger reference policy can lead to improved performance, but only when it is similar to the model being fine-tuned.
Our findings highlight the confounding role of reference policies in DPO and offer insights for best practices, while also identifying open research questions for future studies.
\end{abstract}

\section{Introduction}

Recently, Direct Preference Optimization (DPO )~\cite{rafailov2023direct} has become a widely used training method to align pre-trained large language models with human preferences~\cite{NEURIPS2022_b1efde53}.
DPO offers a significant advantage over standard supervised fine-tuning (SFT) because it learns to distinguish the quality of various candidate outputs, rather than merely relying on a single gold reference.
Specifically, derived from the KL-constrained reinforcement learning (RL) setting, the training objective of DPO implicitly learns a reward model $r_\theta$ given an input $x$ and an output $y$:  $r_\theta(x, y) = \beta \log \frac{p_\theta(y|x)}{p_{\mathrm{ref}}(y|x)}$, 
where $p_{\theta}$ and $p_{\mathrm{ref}}$ are the distributions parameterized by the LLM being fine-tuned and the reference LLM respectively, and $\beta$ controls the strength of the KL divergence regularization applied from the reference LLM.

We argue that the dependency on the reference model/policy $p_{\mathrm{ref}}$,\footnote{We will interchangeably use the terms ``model'' and ``policy'' to reflect the contexts of both LLM fine-tuning and RL.} as demonstrated by the parameterization of $r_\theta$, is an important yet under-explored aspect of DPO.
To begin with, this dependency can lead to a discrepancy between the prediction of $r_\theta(x, y)$, the optimization target, and $p_{\theta}(y|x)$, the learned distribution parameterized by the fine-tuned LLM.
That is, given an input $x$ and a pair of outputs $y_w$ and $y_l$ between which $y_w$ has better quality, even if the reward model correctly predicts that $r_\theta(x, y_w) > r_\theta(x, y_l)$, it does not guarantee $p_{\theta}(y_w|x) > p_{\theta}(y_l|x)$, that the fine-tuned model learns to assign a higher probability to the better output~\cite{chen2024preference}.
Moreover, since this dependency arises from a KL divergence constraint from the reference policy, any deviations from the reference policy will be penalized.
This can create a performance ceiling, as the reference policy is typically instantiated as the SFT model in practice, which is meant to be further improved using DPO.

It thus seems that removing or reducing such constraints might be beneficial.
Recent work~\cite{gorbatovski2024learn} does find that updating the reference policy helps improve DPO's performance.
However, eliminating this dependence can actually lead to performance degradation because of model degeneration~\citep{rafailov2023direct}.
As a result, while related studies have proposed reference-policy-free training methods that have shown superior performance than DPO~\cite{xu2023some, hong2024reference, meng2024simpo}, other forms of regularization are still required in these methods.
For example, ORPO~\cite{hong2024reference} introduces a maximum likelihood estimation (MLE) objective on the positive examples, while SimPO~\cite{meng2024simpo} highlights the importance of length normalization of the sequence log-likelihood.

These findings suggest a complicated relationship between the reference policy and the DPO performance.
Therefore, we aim to further understand the effects and the role of the reference policy in DPO.
To this end, we choose the task of instruction fine-tuning of LLMs~\cite{NEURIPS2022_b1efde53} as the test bed and two series of open-source pre-trained LLMs, Tulu 2 and Mistral~\cite{jiang2023mistral}, to analyze the reference policy in DPO on the AlpacaEval benchmark~\cite{alpaca_eval}.
Specifically, we investigate three main research questions (RQs).

First, 
starting from the standard DPO setting where the KL-constraint regularization is applied using the SFT reference policy,
we explore RQ1: \textbf{what the optimal strength of the KL-constraint should be} (\S\ref{sec:beta}).
Our experiments suggest that the optimal KL-constraint strength is significantly smaller than that used in previous work like Zephyr~\citep{tunstall2023zephyr} and Tulu 2~\cite{ivison2023camels}.
Moreover, we find that DPO is sensitive to this KL-constraint, as setting the constraint slightly smaller than the optimal value can quickly lead to performance degradation or degeneration.
We therefore conduct an in-depth analysis of this sensitivity at both sequence and token levels, which reveals that (1) a small constraint is required for the model to learn to assign higher probabilities to better outputs effectively, but (2) a small constraint also leads to more extreme probability assignments to specific tokens, likely increasing model instability and causing model degeneration.

Having analyzed the standard DPO setting, we then examine the DPO formulation itself and investigate RQ2: \textbf{whether the reference policy is a necessary regularization for DPO} (\S\ref{sec:compare}).
To this end, we compare two alternative reward parameterizations with the DPO reward parameterization $\frac{p_\theta(y|x)}{p_{\mathrm{ref}}(y|x)}$: 
(1) the posterior probability $p_\theta(y|x)$ as the reward, equivalent to a reference-policy-free setting;
(2) the likelihood function $p_\theta(x|y)$ as the reward, equivalent to using the prior language modeling probability $p(y)$ as the reference policy.
Notably, unlike related work~\cite{hong2024reference, meng2024simpo} that replaces the KL-constraint with other types of regularization, we focus on a more controlled comparison under the DPO formulation, which reveals new insights.
Specifically, we provide proof of equivalence between the reference-policy-free preference optimization setting and the Maximum Entropy (MaxEnt) RL setting~\cite{ziebart2008maximum}, allowing a theoretical comparison of these objectives under the RL framework.
Meanwhile, our empirical experiments reveal that the alternative objectives cannot outperform DPO with the optimal KL-constraint strength identified in RQ1. 
Interestingly, this optimal strength is also critical for DPO to outperform reference-policy-free methods like ORPO~\cite{hong2024reference}, reversing their reported superiority.

Given the importance of the reference policy in the DPO objective, the third question (RQ3) we study is: \textbf{whether DPO benefits from a stronger reference policy} (\S\ref{sec:better-ref}).
In detail, rather than the standard practice of using the SFT model to be fine-tuned as the reference model, we employ two more capable LLMs as the reference model.
Our experiments reveal an interesting finding: stronger reference models can indeed offer more benefits than the SFT model, but only when they are compatible with the model being fine-tuned.
Specifically, this compatibility likely stems from the intrinsic similarities of models, such as model architectures or pre-training corpora. 
However, we also find that compatibility does not require the reference model to share the same tokenizer/action space, despite DPO's capability to model dense reward functions which requires a consistent action space as demonstrated by recent work~\cite{rafailov2024r}.

Our work emphasizes the confounding effect of the reference policy in DPO.
Our findings shed light on the best practices for DPO:
(1) when using the SFT model as the reference policy, a smaller yet sufficiently strong constraint typically improves performance;
(2) a stronger reference policy can improve performance but should be compatible with the model to be fine-tuned.
Meanwhile, it highlights open research questions that call for more theoretical analysis of the relationship between the reference policy and the DPO training dynamics.\footnote{Our training scripts, model checkpoints, and datasets are released at \url{https://github.com/yale-nlp/refdpo}.}

\section{Preliminaries}

\subsection{Instruction Fine-tuning}
Instruction fine-tuning aims to align the pre-trained language models with the users' intentions~\cite{NEURIPS2022_b1efde53}.
The frequently used training methods of instruction fine-tuning include standard supervised fine-tuning (SFT), reinforcement learning (RL) methods~\cite{NEURIPS2020_1f89885d, NEURIPS2022_b1efde53} such as Proximal Policy Optimization~\cite{schulman2017proximal}, and reward-based training methods~\cite{rafailov2023direct, yuan2023rrhf, zhao2023slic} such as Direct Preference Optimization~\cite{rafailov2023direct}.
Both RL and reward-based training methods offer advantages over standard SFT by leveraging \textit{human feedback}, typically in the form of human-annotated rewards assigned to various candidates, enabling more effective alignment of LLMs with human preferences.
Therefore, they are widely adopted in the instruction fine-tuning of large language models (LLMs) such as GPT-4~\cite{achiam2023gpt}, Gemini~\cite{team2023gemini}, and Llama-3~\citep{dubey2024llama}.

\subsection{Direct Preference Optimization}

In \citet{rafailov2023direct}, DPO is proposed as an alternative to RL for training LLMs with human feedback.
The DPO objective is derived from the KL-constrained reinforcement learning setting:
\begin{equation}
\label{eq:obj-rl}
\resizebox{\linewidth}{!}{$\max_{p_\theta} \mathbb{E}_{x \sim D, y \sim p_\theta(y|x)}[r(x, y)] - \beta \mathbb{D}_{\mathrm{KL}}[p_\theta(y|x)||p_{\mathrm{ref}}(y|x)],$}
\end{equation}
where $x$ is the input text, $y$ is the output text, $p_{\theta}$ is the probability distribution parameterized by the LLM under fine-tuning with trainable parameters $\theta$, $p_{\mathrm{ref}}$ is a reference distribution which is often instantiated by another LLM, $\beta$ is a parameter controlling the strength of the KL divergence penalty:
\begin{equation}
\label{eq:kl}
\resizebox{0.7\hsize}{!}{%
$\mathbb{D}_{\mathrm{KL}}[p_\theta||p_{\mathrm{ref}}]$ = $\mathbb{E}_{y \sim p_\theta(y|x)}[\log \frac{p_\theta(y|x)}{p_{\mathrm{ref}}(y|x)}]$.
}
\end{equation}
$r(x, y)$ is the reward of $y$ given $x$ under the Bradley-Terry (BT) model~\cite{bradley1952rank}:
\begin{equation}
\label{eq:bt}
\resizebox{0.8\hsize}{!}{%
$p(y_1 \succ y_2|x) = \frac{\exp (r(x, y_1))}{\exp (r(x, y_1)) + \exp(r(x, y_2))} $,
}
\end{equation}
where $p(y_1 \succ y_2|x)$ is the probability of output $y_1$ being better than output $y_2$.
This reward model can be optimized by maximum likelihood estimation: 
\begin{equation}
\label{eq:bt-loss}
\resizebox{0.85\hsize}{!}{%
$\begin{aligned}
\mathcal{L}(r) &= -\mathbb{E}_{(x, y_w, y_l) \sim D}[\log p(y_w \succ y_l|x)] \\
&= -\mathbb{E}_{(x, y_w, y_l) \sim D}[\log\sigma(r(x, y_w) - r(x, y_l))]. \\
\end{aligned}$
}
\end{equation}
Here $x$ is an input in the dataset $D$, $y_l$ and $y_w$ is a pair of outputs between which $y_w$ has better quality, and $\sigma(\cdot)$ is the sigmoid function.

DPO shows that for the aforementioned RL objective (Eq.~\ref{eq:obj-rl}), it is equivalent to optimizing the following objective: learning a reward model parameterized by the model's policy $p_\theta$ (and the reference policy $p_{\mathrm{ref}}$) under the BT model:
\begin{equation}
\label{eq:obj-dpo}
\resizebox{0.85\hsize}{!}{%
$\begin{aligned}
        &\mathcal{L}_{\mathrm{DPO}}(p_\theta;p_{\mathrm{ref}}) 
        = \\ &-\mathbb{E}_{(x, y_w, y_l) \sim D}[\log \sigma (\beta \log \frac{p_\theta(y_w|x)}{p_{\mathrm{ref}}(y_w|x)} - \beta \log \frac{p_\theta(y_l|x)}{p_{\mathrm{ref}}(y_l|x)})],
\end{aligned}$%
}
\end{equation}
where a reward model is implicitly learned:
\begin{equation}
\label{eq:reward}
\resizebox{0.5\hsize}{!}{%
$r_\theta(x,y) = \beta\log\frac{p_\theta(y|x)}{p_{\mathrm{ref}}(y|x)}$.
}
\end{equation}

We note that the reference policy $p_{\mathrm{ref}}$ used in DPO is typically the LLM after supervised fine-tuning, which is the same LLM that will be fine-tuned using DPO.
For simplicity, we refer to this model as the SFT model.

\subsection{Related Work on Preference Learning}
\label{subsec:related-work}
Apart from DPO, various training methods of preference learning have been proposed recently.
For example, IPO~\citep{azar2024general} has been proposed to address the potential limitation of DPO, which tends to overlook the KL-divergence regularization, making it easy to overfit.
On the other hand, many preference learning algorithms are \textit{reference-policy-free}.
Among these, contrastive learning is a major category that interprets the model-predicted probability of an output as a quality score~\cite{liu-etal-2022-brio, zhao2023calibrating, yuan2023rrhf, xu2024contrastive}.
Other related methods also share similar training objectives that are based on the model-predicted probability~\cite{xu2023some, hong2024reference, meng2024simpo}.
These methods have been shown to achieve better or comparable performance to DPO in various settings, without the need for a reference model.
However, we note that additional normalization or regularization is still required for these methods.
For example, an MLE objective on the positive examples is used by SLiC-HF~\cite{zhao2023slic}, CPO~\cite{xu2024contrastive}, ORPO~\cite{hong2024reference}, among others. Meanwhile, SimPO~\cite{meng2024simpo} highlights the importance of length normalization of the sequence log-likelihood in improving method effectiveness.
In this work, we focus on analyzing DPO since it has a more well-established theoretical background while presenting a comparison with reference-free training objectives in \S\ref{sec:compare}.

\section{DPO for Instruction Fine-tuning}
\label{sec:settings}
In this work, we center our analysis in the context of DPO's applications in instruction fine-tuning.
Therefore, here we outline the experimental settings for our analysis in the following sections. 
\paragraph{Datasets}
Following the setup of Zephyr~\citep{tunstall2023zephyr}, we use the UltraFeedback~\cite{cui2023ultrafeedback} dataset, specifically its binarized version,\footnote{The dataset is available at \url{https://huggingface.co/datasets/HuggingFaceH4/ultrafeedback_binarized}.} for DPO fine-tuning.
It contains 64K data examples consisting of an user instruction and a positive-negative output pair scored by GPT-4.\footnote{Further details of datasets are in Appendix~\ref{appendix:datasets}.}

\paragraph{Models}
We mainly use two SFT LLMs as the starting point for DPO fine-tuning from two recent works on instruction fine-tuning.
The first LLM, \texttt{mistral-7b-sft-beta}\footnote{\url{https://huggingface.co/HuggingFaceH4/mistral-7b-sft-beta}} from Zephyr~\citep{tunstall2023zephyr}, is fine-tuned from the Mistral 7B base model~\citep{jiang2023mistral} on the UltraChat~\cite{ding-etal-2023-enhancing} dataset.\footnote{\url{https://huggingface.co/datasets/HuggingFaceH4/ultrachat_200k}}
We will refer to this model as \textbf{\mistral} for simplicity.
The second LLM, \textbf{\texttt{\tulu}}\footnote{\url{https://huggingface.co/allenai/tulu-2-7b}} from Tulu 2~\citep{ivison2023camels}, is fine-tuned from the Llama-2-7B base model~\cite{touvron2023llama} on the TULU v2 SFT mixture dataset.\footnote{\url{https://huggingface.co/datasets/allenai/tulu-v2-sft-mixture}}
We choose these two models for better transparency and reproducibility, since their experimental artifacts are publicly accessible.

\paragraph{Training Settings}
We follow the training recipe proposed in \citet{tunstall2023zephyr} for the experiments, which is also adopted by \citet{ivison2023camels}.
Specifically, the models are fine-tuned with DPO for 3 epochs on the UltraFeedback dataset, with the batch size setting to 32 and linear learning rate
scheduling with warmup.\footnote{More details are in Appendix~\ref{appendix:training}.}
\citet{tunstall2023zephyr} notes that while the models fine-tuned using DPO tend to overfit, this overfitting does not hurt the models' performance, thus they picked the final checkpoint for evaluation.
However, to better understand the effects of the DPO training objective on the final model performance, we instead select the checkpoints based on their loss on the validation set of the UltraFeedback dataset.

\paragraph{Evaluations}
We mainly focus on the evaluations of the LLMs' instruction-following capabilities.
To this end, we use the AlpacaEval2~\cite{alpaca_eval} benchmark for the main evaluation, which consists of 804 test examples for evaluating instruction-following and uses GPT-4 to evaluate the system outputs by comparing them with the outputs generated by GPT-4.\footnote{\texttt{gpt-4-1106-preview} is used to generate the outputs and to perform the pairwise output comparison.}
Notably, its length-controlled version~\cite{dubois2024length}, which mitigates the length bias of GPT-4's evaluations by predicting and canceling it,
achieves a 0.98 Spearman correlation with LMSYS' Chatbot Arena~\cite{chiang2024chatbot} evaluation results collected from crowd-sourced human participants.
Therefore, we mainly report the length-controlled AlpacaEval 2 scores in the following sections.

\section{RQ1: What Is the Optimal KL Constraint Strength for DPO?}
\label{sec:beta}

\begin{table}
    \centering
    \small
\begin{tabular}{lrr}
\toprule
$\beta$ &   \mistral & \tulu \\
\midrule 
$\infty$ (original) & 7.57 &  8.50 \\
0.1 (previous work*) & 13.42  & 9.20  \\
\midrule
0.1 & 14.03 &  9.38\\
0.05 & 13.29 &  9.96 \\
0.02 & 16.06 &  \textbf{10.46} \\
0.01 & \textbf{16.25} &  7.86 \\
0.005 & 12.36  &  [degenerate] \\
\bottomrule
\end{tabular}
%
\caption{Model performance on AlpacaEval2 with varying values of $\beta$ (regularization strength with respect to the reference model). The DPO fine-tuning is started from \mistral and \tulu respectively. *: the DPO fine-tuned models in previous work, Zephyr~\cite{tunstall2023zephyr} and Tulu-2~\cite{ivison2023camels}.}
\label{tab:beta}
\end{table}

The regularization from the reference policy in DPO is introduced by the KL divergence constraint (Eq.~\ref{eq:kl}).
Therefore, we start our analysis of reference policies by investigating the impact of this KL-constraint.
Specifically, in KL-constrained RL and DPO, the weighting coefficient $\beta$ controls the trade-off between maximizing the reward $r$ and minimizing the deviation from the reference policy $p_{\mathrm{ref}}$ (Eq.~\ref{eq:obj-rl}).
Therefore, we vary the value of $\beta$ to understand the impact of this KL constraint.

\subsection{Main Results}
\label{subsec:main-results}
In Table~\ref{tab:beta}, we show the models' performance on AlpacaEval2 that are fine-tuned using DPO from \mistral and \tulu.
Following the original setting of DPO, we also used these two SFT models as the reference models.
We found a similar trend with the two series of fine-tuned models -- \textbf{a smaller KL constraint generally improves performance, until the constraint becomes too small and leads to performance degradation}.\footnote{Additional experiments with a 1.5B LLM are in Appendix~\ref{appendix:rq1}, which demonstrates a similar trend.}

The above observation can be intuitively explained by the need for the fine-tuned model to deviate from the reference model, which is the SFT model targeted for improvement, while maintaining necessary regularization. 
However, we note the algorithm's sensitivity to the strength of the KL constraint: while the studies by Zephyr~\cite{tunstall2023zephyr} and Tulu 2~\cite{ivison2023camels} both set the value of $\beta$ to be $0.1$, the results in Table~\ref{tab:beta} suggest that a smaller optimal value of $\beta$ may be more effective, yet with an increasing risk of performance degradation.
Next, we present further analysis of this KL-constraint strength sensitivity.

\subsection{Ranking Accuracy}
\label{subsec:ranking-accuracy}
Recent work~\cite{chen2024preference} noted a discrepancy between the learned (implicit) reward model and the learned policy model in DPO in terms of their capabilities of correctly ranking candidate outputs according to their quality.
Particularly, the ranking accuracy of the learned reward model (Eq.~\ref{eq:reward}) is usually high, indicating that the better output $y_w$ is assigned a higher reward. i.e., $r_\theta(x, y_w) > r_\theta(x, y_l)$.
However, \citet{chen2024preference} found that the ranking accuracy of the learned policy $p_\theta$ remains low even after DPO fine-tuning.
They partly attributed this to the generally low ranking accuracy of the reference policy $p_{\mathrm{ref}}$, noting that DPO fine-tuning rarely reverses the preference (i.e., $p(y_w|x)$ vs. $p(y_l|x)$) due to the KL-constraint.

We noticed a similar trend in our experiments.
Specifically, Table~\ref{tab:ranking} shows the ranking accuracy of the learned reward model $r_\theta$ and the policy model $p_\theta$ on the validation set of UltraFeedback.
We observe a positive correlation between the strength of the KL constraint ($\beta$) and the ranking accuracy of $p_\theta$ -- \textbf{a sufficiently small $\beta$ is necessary to exceed the $50\%$ ranking accuracy of the random oracle baseline}.
Furthermore, a smaller $\beta$ can lead to a higher policy ranking accuracy, however, as noted in \S\ref{subsec:main-results}, it can cause model degradation.\footnote{\citet{chen2024preference} has also noted that the ranking accuracy and the generation performance of the learned policy are not always positively correlated.}

\begin{table}
    \centering
    \small
\begin{tabular}{@{\extracolsep{1pt}}lcccc@{}}
\toprule
\multirow{2}{*}[-2pt]{$\beta$}         &   \multicolumn{2}{c}{\mistral} & \multicolumn{2}{c}{\tulu} \\
 \cmidrule{2-3} \cmidrule{4-5}
& $r_\theta$ & $p_\theta$ & $r_\theta$ & $p_\theta$ \\
\midrule
$\infty$ (original) &                 0.500 &             0.435 &                0.500 &                0.439 \\
 0.1      &                0.718 &                0.495 &             0.773 &             0.462 \\
 0.05     &                0.744 &                0.513 &             0.778 &             0.483 \\
 0.02     &                0.746 &                0.544 &             0.766 &             0.536 \\
 0.01     &                0.751 &                0.605 &             0.758 &             0.605 \\
 0.005    &                0.766 &                0.704 &             0.750 &             0.680 \\
\bottomrule
\end{tabular}
\caption{Ranking accuracy of the learned reward model $r_\theta(x,y)$ and the policy model $p_\theta(y|x)$ on the Ultrafeedback validation set. 
}
\label{tab:ranking}
\end{table}

\subsection{Token-Level Difference}
\label{subsec:token-level analysis}

\begin{figure}[t!]
    \centering
    \includegraphics[width=0.9\linewidth]{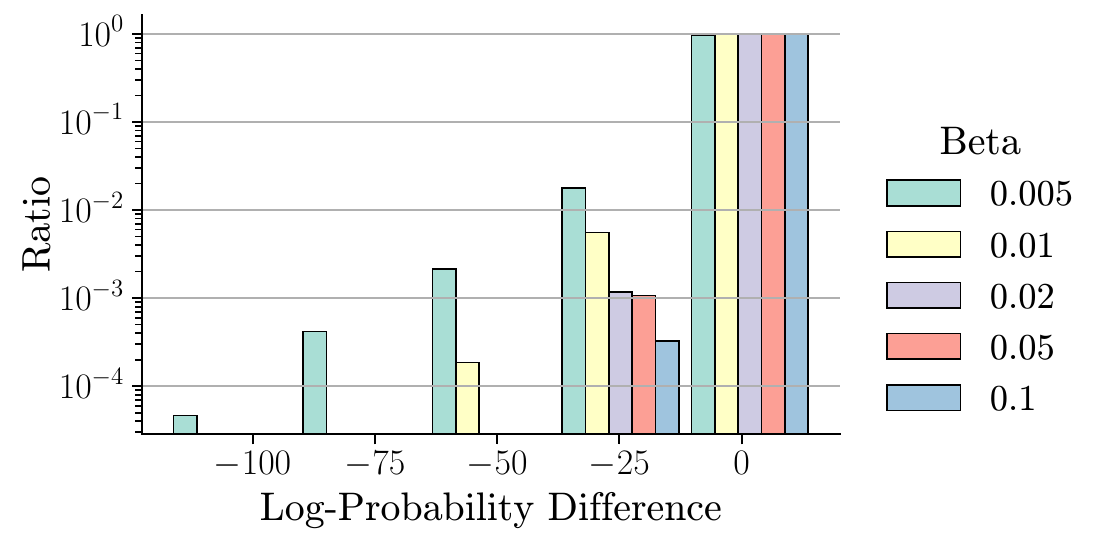}
 \caption{\label{fig:token-weight-avg}The average token log-probability difference (discretized) distribution between the DPO-fine-tuned and reference models (Eq.~\ref{eq:token-weight}).
 The models are fine-tuned from \mistral with different values of $\beta$.
    }
\end{figure}

\citet{rafailov2024r} have recently shown DPO implicitly parameterizes a dense reward function:
\begin{equation}
\resizebox{0.75\hsize}{!}{
$r_\theta(y^{(i)}, s^{(i)}) = \beta \log \frac{p_\theta(y^{(i)}|x, y^{(<i)})}{p_\mathrm{ref}(y^{(i)}|x, y^{(<i)})}$,}
\end{equation}
where $y^{(i)}$ is the $i$-th token (action step) in the output $y$, $y^{(<i)}$ is the prefix of $y$ before $y^{(i)}$.
$s^{(i)}$ is the current ``state'', which is determined by the input $x$ and the prefix $y^{(<i)}$.
Intuitively, as noted by \citet{rafailov2024r}, this formulation implies that DPO can learn token-level credit assignment, which is proportional to the difference between the log probabilities assigned by the trained model and the reference model.
Therefore, we use the statistic, \textit{\textbf{token log-probability difference}}, to study the token-level difference between the trained and reference models:
\begin{equation}
\label{eq:token-weight}
\resizebox{0.75\hsize}{!}{
$\hat{r}_\theta(y^{(i)}, s^{(i)}) = \log \frac{p_\theta(y^{(i)}|x, y^{(<i)})}{p_\mathrm{ref}(y^{(i)}|x, y^{(<i)})}$.}
\end{equation}

Figure~\ref{fig:token-weight-avg} shows the distribution of the average log-probability difference of each token in the model's vocabulary on the UltraFeedback validation set, with models fine-tuned from \mistral.
We note the following:

\noindent (1) Similar to the findings in \citet{rafailov2024r}, we found the DPO fine-tuning leads to a decrease in token probabilities in general. 

\noindent (2) The distribution of token-level log-probability differences is imbalanced, with a small portion of tokens receiving very different probabilities.

\noindent (3) A smaller KL constraint leads to a larger portion of tokens with large probability differences.

\begin{figure}[t!]
    \centering
    \includegraphics[width=0.8\linewidth]{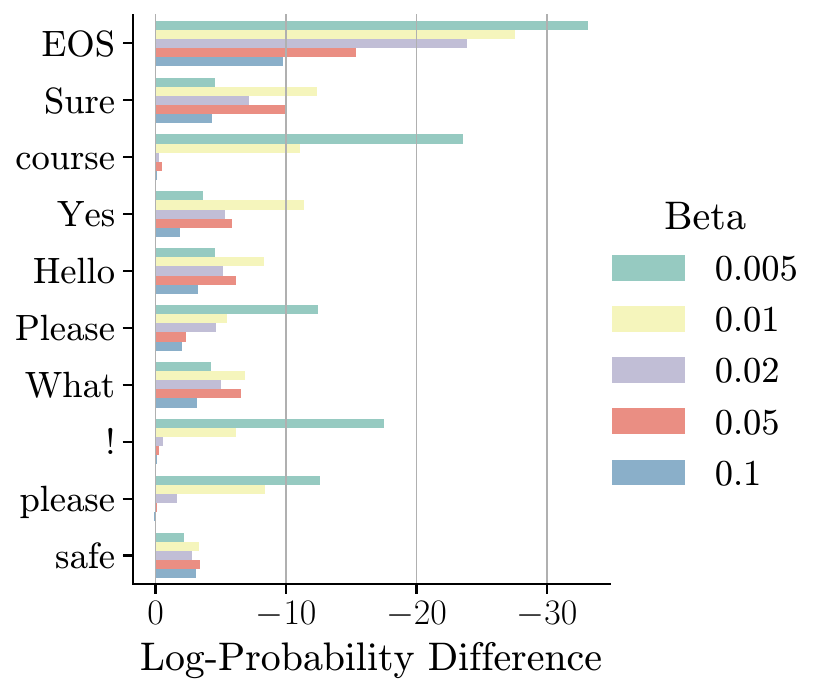}
 \caption{\label{fig:token-weight-most}The tokens most downweighted on average by the models DPO-fine-tuned from \mistral with different values of $\beta$.
    }
\end{figure}

Figure~\ref{fig:token-weight-most} displays the top 10 tokens with the largest probability differences that appear at least 100 times. 
Notably, the end-of-sequence token (EOS) is the most downweighted among them. 
Additionally, a smaller KL-constraint results in a larger difference for the EOS token.
Consequently, we observe that a smaller KL-constraint generally leads to longer outputs, suggesting that the models have learned a preference for output length.

\paragraph{Discussion} 
Our token-level analysis reveals that \textbf{as the strength of the KL-constraint decreases, the DPO-fine-tuned model begins to assign significantly different probabilities to a small subset of specific tokens compared to the reference model}.
The extremeness of these log-probability differences is notable: a difference of $-10$ indicates that the token is downweighted by $e^{10} \approx 22000 $ times on average.
This partially explains the sensitivity to $\beta$, as the increasing extremity of log-probability differences can destabilize the model.
Moreover, it highlights a limitation of the sequence-level KL-constraint: it does not effectively prevent extreme token-level differences.

\section{RQ2: Is Reference Policy a Necessary Regularization for DPO?}
\label{sec:compare}

We now take a step back to examine the necessity of the reference policy as a regularization in DPO.
As discussed in \S\ref{subsec:related-work}, there are already recent studies that proposed \textit{reference-policy-free} algorithms that replace the KL-constraint with other types of regularization.
We are, however, more interested in a closer comparison between DPO and similar training objectives that are either regularization-free or maintain the KL-constraint as the regularization.

\subsection{Training Objectives}
\label{subsec:training-objective}
To this end, we investigate two related training objectives and provide theoretical analyses in the framework established by \citet{rafailov2023direct}: 
optimizing the model using (1) the posterior probability ($p_\theta(y|x)$) and (2) the likelihood function ($p_\theta(x|y)$) as the reward function in the BT model.

\paragraph{Posterior probability as the reward}
Recall that \citet{rafailov2023direct} have shown that the DPO training objective learns a reward model: $r_\theta(x, y)=\beta\log\frac{p_\theta(y|x)}{p_{\mathrm{ref}}(y|x)}$.
An alternative option is to use the posterior probability with a scaling hyperparameter $\beta$ as the reward function, which does not require a reference model:
\begin{equation}
  \tilde{r}_\theta(x, y) = \beta \log p_\theta(y|x).
\end{equation}
The training objective under the BT model (Eq.~\ref{eq:bt-loss}) then becomes
\begin{equation}
\label{eq:obj-prob}
\resizebox{0.85\hsize}{!}{%
$\begin{aligned}
        &\mathcal{L}_{\mathrm{Prob}}(p_\theta) 
        = \\ &-\mathbb{E}_{(x, y_w, y_l) \sim D}[\log \sigma (\beta \log p_\theta(y_w|x) - \beta \log p_\theta(y_l|x))].
\end{aligned}$%
}
\end{equation}
This objective has been investigated in recent work~\cite{xu2024contrastive} for fine-tuning LLMs for machine translation, but in \citet{xu2024contrastive} it is used together with an MLE loss on the positive examples as a necessary regularization. 

Following the analytical framework in \citet{rafailov2023direct}, we note that optimizing $L_\mathrm{Prob}(p_\theta)$ is equivalent to optimizing the policy model under the following RL objective:
\begin{equation}
\label{eq:obj-maxent-rl}
\resizebox{0.85\hsize}{!}{$\max_{p_\theta} \mathbb{E}_{x \sim D, y \sim p_\theta(y|x)}[r(x, y)] + \beta \mathrm{H}[p_\theta(y|x)]$,}
\end{equation}
where $\mathrm{H}$ is the entropy function: 
\begin{equation}
    \mathrm{H}[p_\theta(y|x)] = \mathbb{E}_{y \sim p_\theta(y|x)}[- \log p_\theta(y|x)].
\end{equation}
The proof for the equivalence of Eq.~\ref{eq:obj-prob} and Eq.~\ref{eq:obj-maxent-rl} is in Appendix~\ref{appendix:prob}, with further analyses showing that Eq.~\ref{eq:obj-prob} shares similar properties as DPO.

We note that Eq.~\ref{eq:obj-maxent-rl} resembles the Maximum Entropy (MaxEnt) RL setting~\cite{ziebart2008maximum, eysenbach2021maximum} -- apart from optimizing for a higher reward, this objective discourages the model from becoming overly deterministic.

\paragraph{Likelihood function as the reward}
Apart from the posterior probability, another candidate reward function is the likelihood function $p(x|y)$.
Specifically, by Bayes' theorem, we have
\begin{equation}
    p(y|x) \propto p(x|y) p(y).
\end{equation}

Under the task of instruction-following, $p(x|y)$ depicts how likely the instruction $x$ is related to the model response $y$.
With the above factorization, we can derive the learning objective with $p_\theta(x|y)$ as the reward function:
\begin{equation}
\resizebox{0.4\hsize}{!}{$p_\theta(x|y) = \frac{p_\theta(y|x)p(x)}{p_\mathrm{LM}(y)}.$}
\end{equation}
Here, 
we use a fixed LLM to parameterize the prior probability $p(y)$, which we label $p_\mathrm{LM}(y)$.\footnote{We use the SFT model in this work.}
We will show later that $p(x)$ does not need an explicit parameterization.

Using $p_\theta(x|y)$ as the reward function with a scaling factor $\beta$, the objective under the BT model (Eq.~\ref{eq:bt-loss}) becomes
\begin{equation}
\label{eq:obj-likelihood}
\resizebox{0.86\hsize}{!}{%
$
\begin{aligned}
     & \mathcal{L}_{\mathrm{Likelihood}}(p_\theta)\\
     &= -\mathbb{E}_{(x, y_w, y_l) \sim D}[\log \sigma (\beta \log p_\theta(x|y_w) - \beta \log p_\theta(x|y_l))] \\
        &= -\mathbb{E}_{(x, y_w, y_l) \sim D}[\log \frac{(\frac{p_\theta(y_w|x)}{p_\mathrm{LM}(y_w)})^\beta}{(\frac{p_\theta(y_w|x)}{p_\mathrm{LM}(y_w)})^\beta + (\frac{p_\theta(y_l|x)}{p_\mathrm{LM}(y_l)})^\beta}]
\end{aligned}  
$%
}
\end{equation}
Eq.~\ref{eq:obj-likelihood} is equivalent to using $\frac{p_\theta(y|x)}{p_{\mathrm{LM}}(y)}$ as the reward function, which has an intuitive interpretation -- the input-dependent quality score of $y$ is its conditional probability given $x$, $p_\theta(y|x)$, normalized by its unconditional probability, $p_\mathrm{LM}(y)$.

Furthermore, under the analytical framework of DPO, we note that Eq.~\ref{eq:obj-likelihood} is equivalent to the KL-constrained RL setting (Eq.~\ref{eq:obj-rl}) with the prior probability $p_\mathrm{LM}(y)$ as the reference policy.
That is,
\begin{equation}
\label{eq:obj-likelihood-rl}
\resizebox{0.85\hsize}{!}{$\max_{p_\theta} \mathbb{E}_{x \sim D, y \sim p_\theta(y|x)}[r(x, y)] - \beta \mathbb{D}_{\mathrm{KL}}[p_\theta(y|x)||p_{\mathrm{LM}}(y)].$}
\end{equation}

\subsection{Result Analysis}

\begin{table}
    \centering
    \small
\addtolength{\tabcolsep}{-2pt} 
\begin{tabular}{lrrrr}
\toprule
\textbf{Method} & \textbf{Score}      &  \textbf{Accuracy} & \textbf{Length} & $\boldsymbol{\beta}$ \\
\midrule
original (SFT) & 7.57 & 0.435 & 205.0 & - \\
DPO (Eq.~\ref{eq:obj-dpo}) & 16.25 & 0.605 &  359.4 & 0.01\\
probability (Eq.~\ref{eq:obj-prob}) & 12.84  & 0.697 &  744.5 & 100.0 \\
likelihood (Eq.~\ref{eq:obj-likelihood}) & 13.63 & 0.602 & 389.3 & 0.01\\
\bottomrule
\end{tabular}
\addtolength{\tabcolsep}{+2pt} 
\caption{Performance comparison of different training objectives for fine-tuning \mistral.
The best performance with the optimal $\beta$ is reported.
The AlpacaEval2 \textbf{score}, the \textbf{ranking accuracy} of the model's probability, and the output \textbf{length} are reported.}
\label{tab:policies}
\end{table}

We use \mistral as the base model for fine-tuning to compare DPO with the other two reward parameterizations discussed above.
Table~\ref{tab:policies} summarizes the performance of different algorithms with the optimal value of $\beta$.
We found that

\noindent (1) All fine-tuning methods improve the performance of the original SFT model;

\noindent (2) Using posterior probability as the reward function (Eq.~\ref{eq:obj-prob}) 
needs a large optimal value of $\beta$, likely because higher entropy helps prevent degeneration.

\noindent (3) DPO outperforms the alternative training objectives. 
Notably, DPO also outperforms the reference-policy-free ORPO method~\cite{hong2024reference}, which scored 14.7 on the AlpacaEval2 dataset under a comparable setting.\footnote{The result is available at \url{https://tatsu-lab.github.io/alpaca_eval/}.}
This is despite ORPO's superior performance to Zephyr~\cite{tunstall2023zephyr}, which uses a suboptimal DPO configuration with $\beta$ set to $0.1$, as shown in Table~\ref{tab:beta}.

\begin{figure}[t!]
    \centering
    \includegraphics[width=0.9\linewidth]{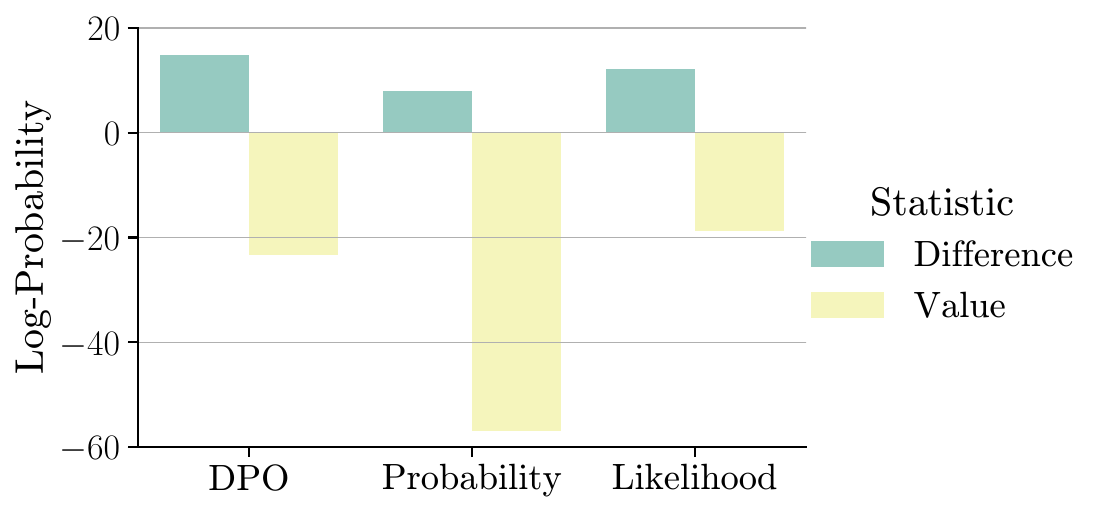}
 \caption{\label{fig:eos-study}Average log-probability (\textbf{value}) and the average log-probability \textbf{difference} between positive and negative examples of the EOS token, as assigned by models fine-tuned with different training objectives.
    }
\end{figure}
\paragraph{Case Study}
To further understand the difference in these training objectives, we present a case study with the EOS token, as we have found in \S\ref{subsec:token-level analysis} that DPO training can significantly alter model behavior regarding the EOS token.
Therefore, here we provide a further investigation.
Specifically, we calculate two statistics:
(1) the average log-probability assigned to the EOS token;
(2) the average difference in the log-probability assigned to the EOS token between positive ($y_w$) and negative ($y_l$) examples, illustrating how the model behaves differently in these two scenarios with the EOS token.

Figure~\ref{fig:eos-study} presents these two statistics of models fine-tuned with different reward function parameterizations (averaged over different values of $\beta$).
We note the difference between DPO and the fine-tuning method that uses posterior probability as the reward function: 
the former achieves a larger distinction between positive and negative examples in terms of the probability assignment to the EOS token, while the latter shows a smaller distinction, despite generally assigning more extreme probabilities to the EOS token, which leads to significantly longer outputs.
This suggests that \textbf{the KL-constraint from the reference policy in DPO helps to stabilize the model behavior}.
We present a qualitative case study of the outputs generated with different training objectives in Appendix~\ref{appendix:case}.

\section{RQ3: Does DPO Benefit from Stronger Reference Policies?}
\label{sec:better-ref}

In DPO, the reference policy is instantiated with the SFT model for further fine-tuning. 
Previous sections have highlighted the significance of reference policies in DPO.
We now explore whether stronger reference policies enhance DPO.

\subsection{Experimental Settings}
We choose two LLMs as the stronger reference policies in DPO for the fine-tuning of \mistral and \tulu:
(1) \texttt{mistral-7b-instruct-v0.2},\footnote{\url{https://huggingface.co/mistralai/Mistral-7B-Instruct-v0.2}}
(2) \texttt{llama-3-70b-instruct}.\footnote{\url{https://huggingface.co/meta-llama/Meta-Llama-3-70B-Instruct}}
These two models are more capable than the base models, achieving length-controlled AlpacaEval2 scores of 20.55 and 34.40 respectively.
We refer to them as \mistralref and \llama for simplicity.

\subsection{Result Analysis}

Table~\ref{tab:tulu-refs} displays model performance after DPO fine-tuning across different $\beta$ values and reference models.
For easier comparison, Table~\ref{tab:tulu-refs-compare} lists the optimal performance using different reference models.
We note the following:

\noindent (1) \textbf{A stronger reference policy in DPO fine-tuning can improve DPO's effectiveness}.
Specifically, the model fine-tuned from \mistral performs best with \mistralref as the reference, and the model fine-tuned from \tulu performs best with \llama as the refererence.

\noindent (2) However, \textbf{a stronger reference policy in DPO fine-tuning does not always lead to better performance}.
Notably, DPO fine-tuning of \tulu with \mistralref as the reference policy did not improve \tulu's original performance.

\noindent (3) \textbf{The optimal KL constraint strength is larger with a stronger, suitable reference policy.} 
In particular, as shown in Table~\ref{tab:tulu-refs}, the optimal value of $\beta$ for \mistral and \tulu is 1.0 with their respective suitable reference policies.
Conversely, when the SFT model is used as the reference policy, the optimal $\beta$ is 0.01 or 0.02, as shown in Table~\ref{tab:beta}.\footnote{Additional experiments with a DPO-fined LLM as the reference policy are in Appendix~\ref{appendix:rq3}.}

\begin{table}
    \centering
    \small
\addtolength{\tabcolsep}{-2pt} 
\begin{tabular}{@{\extracolsep{1pt}}lrrrr@{}}
\toprule
\multirow{2}{*}[-2pt]{$\beta$}         &   \multicolumn{2}{c}{\mistral} & \multicolumn{2}{c}{\tulu} \\
 \cmidrule{2-3} \cmidrule{4-5}
&  \texttt{mistral-v0.2}  &  \texttt{llama3}  &   \texttt{mistral-v0.2}  &  \texttt{llama3}  \\
\midrule
10.0 & 18.74  &  13.29  & 7.61  &   9.79  \\
1.00 & \textbf{20.25}  &   9.59 & \textbf{7.85}  &   \textbf{11.17}\\
0.10 & 19.58 & 10.99 & [degenerate] & 10.31\\
0.01 & 17.18 & \textbf{15.37}  & [degenerate] & 9.16 \\
0.005 & 15.34 & 11.70  & [degenerate] & 3.29 \\
\bottomrule
\end{tabular}
\addtolength{\tabcolsep}{+2pt} 
\caption{Model performance on AlpacaEval2 with varying values of $\beta$ fine-tuned from \mistral and \tulu. Two reference models, \mistralref and \llama are used.}
\label{tab:tulu-refs}
\end{table}

\begin{table}
    \centering
    \small
\addtolength{\tabcolsep}{-2pt} 
\begin{tabular}{lrrrr}
\toprule
& original         &   self & \texttt{mistral-v0.2} &   \texttt{llama3}  \\
\midrule
\texttt{mistral-7b} & 7.57 & 16.25 & 20.25 & 15.37 \\
\tulu & 8.50 & 10.46 & 7.85 & 11.17 \\
\bottomrule
\end{tabular}
\addtolength{\tabcolsep}{+2pt} 
\caption{Model performance on AlpacaEval2.
The best performance across different values of $\beta$ with different reference models is compared, including the SFT model itself (self), \mistralref, and \llama.}
\label{tab:tulu-refs-compare}
\end{table}

\subsection{Discussion}
\label{sec:discussion}
Our investigation reveals while a stronger reference model can be helpful in DPO, it is not always the case.
One possible explanation is that the reference model should be similar enough to the model under training to be compatible -- \mistral performs better with \mistralref as the reference since they are fine-tuned from the same base model, while \tulu performs better with \llama because \tulu is fine-tuned from \texttt{llama-2-7b}.

Regarding model similarity and the choice of the reference policy, we acknowledge the need for analytical tools and call for further research in this area.
The KL divergence is a potential method, but estimating it is challenging due to the large sampling space, leading to high variance. 

We note an interesting aspect in the fine-tuning of \tulu using \llama as the reference policy.
Specifically, these models use different tokenizers, meaning that they do not share the same token-level action space.
The fact that this combination yields further improvement suggests that action spaces do not always need to be aligned in DPO, despite related work proving that DPO can implicitly parameterize a dense reward function, which requires a consistent action space~\cite{rafailov2024r}.

\section{Conclusion}
In this work, we studied the effects of reference policies in DPO, which can stabilize the training while introducing a potential performance upperbound.
We found that DPO's performance is sensitive to the strength of constraints from the reference policy and demonstrated the benefits of using reference policies as a regularization in DPO.
We also showed that a stronger reference policy can enhance DPO performance, but only when the reference policy is compatible.
Our findings highlight the confounding role of reference policies in DPO, providing empirical insights and emphasizing the need for a more in-depth theoretical analysis.
We call for future work to provide more theoretical and empirical guidelines for choosing the constraint strength from the reference policy, as well as understanding the similarity and compatibility between the model to be trained and the reference model.

\section*{Limitations}
\paragraph{Evaluations} 
We mainly use the AlpacaEval2 benchmark to evaluate the instruction-following capabilities of LLMs in this work, on which the LLM-based evaluation achieves a high correlation with human evaluations~\cite{dubois2024length}.
The test examples on this benchmark cover a wide range of instruction types~\cite{dubois2024alpacafarm}, which enhances the generalizability of the evaluation results. 
However, we acknowledge that the LLM-based automatic evaluation has limitations, such as low self-consistency rates in their predictions and a preference for the LLMs' own outputs~\cite{liu-etal-2023-g, wang2023large, panickssery2024llm}. 
Therefore, while we did not perform human evaluations because the large number of analyses we conducted makes it cost-ineffective, we acknowledge its importance in terms of achieving more faithful evaluation results.

\paragraph{Model Scale}
The LLMs we studied in this work primarily consist of around 7 billion parameters. Due to computing resource constraints, we did not investigate LLMs with more parameters. Therefore, whether our observations are applicable on a larger scale remains to be investigated.

\section*{Acknowledgements} 
We thank Hamish Ivison for the helpful suggestions regarding the training details for DPO.
We are grateful for the TPU compute support provided by the Google TRC program and for the OpenAI API credits support provided by the OpenAI's Researcher Access Program.

\bibliography{anthology,custom}

\appendix

\section{Additional Experimental Details}
\label{appendix:exp}

\subsection{Datasets Details}
\label{appendix:datasets}
Here we outline additional details regarding the datasets we used (\S\ref{sec:settings}).
Following the setup of Zephyr~\citep{tunstall2023zephyr}, we use the UltraFeedback~\cite{cui2023ultrafeedback} dataset, specifically its binarized version,\footnote{The dataset is available at \url{https://huggingface.co/datasets/HuggingFaceH4/ultrafeedback_binarized}.} for DPO fine-tuning.
The original UltraFeedback contains 64K data examples consisting of a user instruction and 4 candidate outputs generated by different LLMs, which are scored by GPT-4.
\citet{tunstall2023zephyr} binarized UltraFeedback to modify the data format for DPO fine-tuning, by constructing a positive-negative example pair out of the 4 candidate outputs.
Specifically, the output with the highest GPT-4 score is selected as the positive example $y_w$ used in DPO fine-tuning (Eq.~\ref{eq:obj-dpo}), where the negative example $y_l$ is randomly sampled from the rest 3 candidate outputs.
We note that since UltraFeedback contains LLM-generated, GPT-4 graded outputs, the instruction fine-tuning conducted on UltraFeedback is a distillation setting from more capable LLMs, instead of learning directly from human feedback.
UltraFeedback is released under the MIT license.

\subsection{Additional Training Details}
\label{appendix:training}
Here we discuss additional training details in \S\ref{sec:settings}.
We use a linear learning rate scheduler during model training, with 10\% of the steps allocated for learning rate warmup and the remainder for linearly decreasing the rate.
We perform checkpoint selection based on the model validation loss on the UltraFeedback validation set, with a checkpointing interval of 500 steps.
We conduct model training on 8 NVIDIA RTX 6000 Ada cards with 40GB memory each.
The 3-epoch training takes around 9 hours to finish.

\section{Additional Experiments for RQ1}
\label{appendix:rq1}

\begin{table}
    \centering
    \small
\begin{tabular}{lr}
\toprule
$\beta$ &   AlpacaEval2 Score \\
\midrule 
SFT & 4.82 \\
\midrule
0.1 & 4.75 \\
0.05 & 5.00 \\
0.02 & \textbf{5.89}  \\
0.01 & 4.91  \\
0.005 & 5.52 \\
0.001 & [degenerate] \\
\bottomrule
\end{tabular}
\caption{Model performance on AlpacaEval2 with varying values of $\beta$ (regularization strength with respect to the reference model). The DPO fine-tuning is started from an SFT checkpoint fine-tuned from qwen2-1.5b.}
\label{tab:beta-qwen}
\end{table}

In \S\ref{sec:beta}, we study our RQ1: what the optimal strength of the KL-constraint should be, using two series of LLMs with 7B parameters.
Here, we present additional results with a 1.5B pre-trained LLM, \textbf{qwen2-1.5b}~\cite{yang2024qwen2}.\footnote{\url{https://huggingface.co/Qwen/Qwen2-1.5B}}.
The training setting for qwen2-1.5b is the same as the ones used in \S\ref{sec:beta}, except that we trained the SFT model using the UltraChat dataset.

Table~\ref{tab:beta-qwen} presents experimental results, showing a similar trend as the experiments in \S\ref{subsec:main-results} -- a small KL-constraint helps the model to achieve the optimal performance.
Notably, the optimal strength of the KL-constraint identified is the same as the one for \tulu.

\section{Extending DPO: Posterior Probability as Reward Function}
\label{appendix:prob}

Following the analytical framework in \citet{rafailov2023direct}, we now prove the equivalence of Eq.~\ref{eq:obj-prob} and Eq.~\ref{eq:obj-maxent-rl} as previously discussed in \S\ref{subsec:training-objective}.

\subsection{Optimum of RL with Maximum Entropy}
\label{appendix:a1}
We will first derive the optimal policy, $p^*$, for the RL with Maximum Entropy setting (Eq.~\ref{eq:obj-maxent-rl}):
\begin{equation}
\resizebox{0.85\hsize}{!}{$\max_{p} \mathbb{E}_{x \sim D, y \sim p(y|x)}[r(x, y)] + \beta \mathrm{H}[p(y|x)]$.}
\end{equation}
Given any reward function $r$, we have
\begin{equation}
\label{eq:probability-proof}
\resizebox{0.85\hsize}{!}{$\begin{aligned}
     & \max_{p} \mathbb{E}_{x \sim D, y \sim p(y|x)}[r(x, y)] + \beta \mathrm{H}[p(y|x)] \\
  = &\max_{p} \mathbb{E}_{x \sim D, y \sim p(y|x)}[r(x, y)] - \beta \log p(y|x)] \\
  = &\min_{p} \mathbb{E}_{x \sim D, y \sim p(y|x)}[\log p(y|x) - \frac{1}{\beta}r(x,y)] \\
  = &\min_{p} \mathbb{E}_{x \sim D, y \sim p(y|x)}[\log \frac{p(y|x) }{\frac{1}{Z(x)}\exp(\frac{1}{\beta}r(x,y))}- \log Z(x)],
\end{aligned}$}
\end{equation}
where $Z(x)$ is the partition function:
\begin{equation}
\resizebox{0.5\hsize}{!}{$Z(x) = \sum_y \exp (\frac{1}{\beta}r(x,y))$.}
\end{equation}
As $Z(x)$ is only a function of $x$ and does not depend on $p$, we can define the following probability distribution:
\begin{equation}
\resizebox{0.5\hsize}{!}{$p^*(y|x) = \frac{1}{Z(x)}\exp (\frac{1}{\beta}r(x,y))$.}
\end{equation}
Since $Z(x)$ is not a function of $y$, we can re-organize the final objective of Eq.~\ref{eq:probability-proof} as:
\begin{equation}
\label{eq:probability-proof-cont}
\resizebox{0.85\hsize}{!}{$\begin{aligned}
     & \min_{p} \mathbb{E}_{x \sim D}[\mathbb{E}_{ y \sim p(y|x)}[\log\frac{p(y|x)}{p^*(y|x)}] - \log Z(x)] \\
     = & \min_{p} \mathbb{E}_{x \sim D}[\mathbb{D}_{\mathrm{KL}}[p(y|x)||p^*(y|x)] - \log Z(x)]
\end{aligned}$}
\end{equation}
Since $Z(x)$ does not depend on $p(y|x)$, the final objective of Eq.~\ref{eq:probability-proof-cont} is equivalent to 
\begin{equation}
\label{eq:probability-proof-cont-cont}
\resizebox{0.6\hsize}{!}{$
  \min_{p} \mathbb{E}_{x \sim D}[\mathbb{D}_{\mathrm{KL}}[p(y|x)||p^*(y|x)] ]
$}
\end{equation}
By Gibbs' inequality, the KL-divergence achieves the minimum of 0 if and only if the two distributions are identical.
Therefore, we have the optimal solution of Eq.~\ref{eq:probability-proof-cont-cont} (and Eq.~\ref{eq:obj-maxent-rl}) as 
\begin{equation}
\label{eq:probability-proof-final}
\resizebox{0.8\hsize}{!}{$
 p(y|x) = p^*(y|x) = \frac{1}{Z(x)}\exp (\frac{1}{\beta}r(x,y)).
$}
\end{equation}

\subsection{Extending DPO Objective for RL with Maximum Entropy}
We can now express the reward function $r(x, y)$ under the RL with maximum entropy using the optimal policy $p^*(y|x)$ by rearranging Eq.~\ref{eq:probability-proof-final}:
\begin{equation}
\label{eq:probability-reward}
\resizebox{0.7\hsize}{!}{$
 r^*(x, y) = \beta \log p^*(y|x) + \beta \log Z(x)
$}
\end{equation}
Therefore, we know that the optimal policy $p^*$ under the BT model (Eq.~\ref{eq:bt}) satisfies the following preference model:
\begin{equation}
\label{eq:optimal-prob}
\resizebox{1.0\hsize}{!}{$
\begin{aligned}
     &p^*(y_1 \succ y_2|x) \\
     =& \frac{\exp (\beta \log p^*(y_1|x) + \beta \log Z(x))}{\exp (\beta \log p^*(y_1|x) + \beta \log Z(x)) + \exp(\beta \log p^*(y_2|x) + \beta \log Z(x))} \\
     = &\frac{1}{1 + \exp (\beta\log p^*(y_1|x) - \beta \log p^*(y_2|x) )} \\
     =& \sigma (\beta \log p^*(y_1|x) - \beta \log p^*(y_2|x)).
\end{aligned}
$}
\end{equation}
We can now derive the maximum likelihood estimation objective based on Eq.~\ref{eq:optimal-prob} for the model $p_\theta$:
\begin{equation}
\label{eq:obj-prob-v2}
\resizebox{1.0\hsize}{!}{%
$\begin{aligned}
       \min_{p_\theta} &-\mathbb{E}_{(x, y_w, y_l) \sim D}[\log \sigma (\beta \log p_\theta(y_w|x) - \beta \log p_\theta(y_l|x))]
\end{aligned}$%
}
\end{equation}
This is exactly the objective function we defined in \S\ref{subsec:training-objective} that uses the posterior probability as the reward function (Eq.~\ref{eq:obj-prob}).

\subsection{Further Theoretical Analysis}
We continue to follow the analytical framework of DPO to discuss the theoretical properties of the training objective defined in Eq.~\ref{eq:obj-prob-v2} (and Eq.~\ref{eq:obj-prob}).
Specifically, as we discussed in \S\ref{subsec:training-objective}, the training objective of Eq.~\ref{eq:obj-prob-v2} is equivalent to a BT model with the reparameterization: 
\begin{equation}
  r_\theta(x, y) = \beta \log p_\theta(y|x).
\end{equation}
Following the proof framework in \citet{rafailov2023direct}, we will now prove:

\noindent \textit{all reward classes under the BT model can be represented with the reparameterization $r(x, y) = \beta \log p(y|x)$ for some model $p(y|x)$.}

\paragraph{Preparation}
Our proof uses the definition of the equivalent class of reward functions given by \citet{rafailov2023direct}:

\noindent
    ``\textit{Two reward functions $r(x,y)$ and $r'(x,y)$ are equivalent iff $r(x,y) - r'(x,y) = f(x)$ for some function $f$.}''

We also note a lemma given by \citet{rafailov2023direct}.

\noindent
``\textit{Under the Bradley-Terry preference framework, two reward functions from the same class induce the same preference distribution.}''

We now provide another lemma as an extension of a similar lemma in \citet{rafailov2023direct}:

\noindent \textit{Two reward functions from the same equivalent class induce the same optimal policy under the RL with Maximum Entropy setting}.

\noindent \textit{Proof}. Given two reward functions from the same class, such that $r'(x,y) = r(x,y) + f(x)$, from Eq.~\ref{eq:probability-proof-final} we know that $r'$ induces an optimal policy $p_{r'}$:
\begin{equation}
\resizebox{1.0\hsize}{!}{$
\begin{aligned}
p_{r'}(y|x) &=  \frac{1}{\sum_y \exp (\frac{1}{\beta}r'(x,y))}\exp (\frac{1}{\beta}r'(x,y)) \\
&=  \frac{1}{\sum_y \exp (\frac{1}{\beta}(r(x,y) + f(x)))}\exp (\frac{1}{\beta}(r(x,y) + f(x))) \\
&=  \frac{1}{\exp(\frac{1}{\beta}f(x))\sum_y \exp (\frac{1}{\beta}r(x,y))}\exp (\frac{1}{\beta}r(x,y)) \exp(\frac{1}{\beta}f(x)) \\
&=  \frac{1}{\sum_y \exp (\frac{1}{\beta}r(x,y))}\exp (\frac{1}{\beta}r(x,y))\\
&= p_r(y|x).
\end{aligned}
$}
\end{equation}
\textit{Proof end.}
As \citet{rafailov2023direct} suggested, this lemma indicates that for a certain reward equivalence class, any arbitrary reward function within the class will induce the same optimal policy.

\paragraph{Proof} For any reward function $r(x,y)$, from Appendix~\ref{appendix:a1}, we know it induces the following optimal policy according to Eq.~\ref{eq:probability-proof-final}:
\begin{equation}
\resizebox{0.6\hsize}{!}{$
 p_r(y|x) = \frac{1}{Z(x)}\exp (\frac{1}{\beta}r(x,y)).
$}
\end{equation}
Therefore, we have
\begin{equation}
\resizebox{0.7\hsize}{!}{$
 r(x, y) = \beta \log p_r(y|x) + \beta \log Z(x),
$}
\end{equation}
where $Z(x) = \sum_y \exp (\frac{1}{\beta}r(x,y))$.
We now define a projection $f$:
\begin{equation}
\resizebox{0.7\hsize}{!}{$
f(r;\beta) = r(x,y) - \beta\log\sum_y\exp{(\frac{1}{\beta}r(x,y))},
$}
\end{equation}
which projects $r$ into a new reward function:
\begin{equation}
\resizebox{0.7\hsize}{!}{$
\begin{aligned}
r'(x,y) &= f(r;\beta)(x,y)  \\
&= r(x,y) - \beta\log\sum_y\exp{(\frac{1}{\beta}r(x,y))} \\
&= \beta \log p_r(x,y).
\end{aligned}
$}
\end{equation}
We can observe that $r'$ is within the equivalence class of $r$ since the projector $f$ only depends on $x$.
Our proof is therefore complete.

\section{Output Examples from Different Training Objectives}
\label{appendix:case}

We show a specific output example for each of the training objectives investigated in \S\ref{sec:compare}.
Compared with the output generated from the DPO objective in Table~\ref{tab:output-dpo}, the posterior probability  objective (Eq.~\ref{eq:obj-prob}) produces longer outputs with more complex structures that appear overly verbose in Table~\ref{tab:output-probability}.
Conversely, the likelihood function objective (Eq.~\ref{eq:obj-likelihood}) results in less detailed outputs, as shown in Table~\ref{tab:output-likelihood}.

\section{Additional Experiments for RQ3}
\label{appendix:rq3}

\begin{table}
    \centering
    \small
\begin{tabular}{lr}
\toprule
$\beta$ &   AlpacaEval2 Score \\
\midrule 
reference ($\beta$ = 0.01) & 16.24 \\
\midrule
10.0 & 15.20 \\
1.00 & \textbf{16.75} \\
0.10 & 14.99  \\
0.01 & 15.07  \\
0.005 & 14.80 \\
\bottomrule
\end{tabular}
\caption{Model performance on AlpacaEval2 with varying values of $\beta$.
The DPO fine-tuning is started from \mistral, and the reference model is the checkpoint that is already fine-tuned using DPO from \mistral.}
\label{tab:rq3-iter}
\end{table}

In \S\ref{sec:better-ref}, we study our RQ3: whether DPO benefits from a stronger reference policy, using two stronger reference LLMs.
Here, we present additional experiments with another option of a stronger LLM -- an LLM that has already been trained using DPO.
Specifically, we use the optimal DPO checkpoint fine-tuned from \mistral as the reference model, which is trained with the strength of the KL-constraint set to $0.01$ (\S\ref{subsec:main-results}).
This model is then used to fine-tune \mistral again.

The experimental results are presented in Table~\ref{tab:rq3-iter}, aligned with our observations in \S\ref{sec:better-ref} -- the optimal strength of the KL-constraint is larger when a stronger, suitable reference policy is used.
Specifically, the optimal value of $\beta$ identified is $1.0$, much larger than the optimal value identified in \S\ref{subsec:main-results}, $0.01$.
Interestingly, the optimal fine-tuned model under this setting outperforms the reference policy used, further demonstrating the benefit of stronger reference policies.

\begin{table*}[ht]
  \centering
  \footnotesize
    \begin{tabular}{p{50em}}
    \toprule
   \textbf{Instruction} \\
I like to host guests at my home from time to time, and I am gathering recipes of different dishes and drinks to keep things interesting. I am interested in trying some Somali dishes. Can you give me a recipe for Canjeero?
\\
\midrule
\textbf{Output} \\
Certainly! Here's a traditional recipe for Canjeero (also known as Cannoli or Xaaj maanta in Somali), a popular Somali flatbread made with teff flour or a mix of teff and wheat flour.\newline \newline Ingredients:\newline - 2 cups teff flour (you can find this at specialty stores or online)\newline - 1 cup wheat flour (or more teff flour if you prefer a stronger teff flavor)\newline - 2 tsp salt\newline - 1 package or 2 1/4 tsp (7g) active dry yeast\newline - 3 cups warm water (around 110°F)\newline \newline Instructions:\newline \newline 1. In a large mixing bowl, combine the teff flour, wheat flour, and salt. Mix well.\newline \newline 2. In a separate bowl, dissolve the yeast in the warm water. Let it sit for 5 minutes until it becomes frothy.\newline \newline 3. Pour the yeast mixture into the flour mixture. Mix well until a smooth, sticky dough forms.\newline \newline 4. Let the dough rest in a warm place for 1-2 hours until it doubles in size.\newline \newline 5. When the dough has risen, divide it into six equal portions. Shape each portion into a ball.\newline \newline 6. Preheat a flat griddle or a non-stick pan over medium heat.\newline \newline 7. Take one ball of the dough and flatten it into a circle, about 1/4 inch thick.\newline \newline 8. Place the flattened dough on the hot griddle or pan. Cook for 2-3 minutes until the bottom is golden brown. Flip the Canjeero and cook for another 1-2 minutes until the other side is also golden brown.\newline \newline 9. Remove the Canjeero from the griddle or pan and keep it warm and moist by covering it with a clean towel or aluminum foil.\newline \newline 10. Repeat steps 7-9 with the remaining dough portions.\newline \newline Canjeero is often served with a stew or soup. Enjoy!
\\
    \bottomrule
    \end{tabular}%
\caption{Output example with the DPO objective (Eq.~\ref{eq:obj-dpo}).}
  \label{tab:output-dpo}
\end{table*}%

\begin{table*}[ht]
  \centering
  \footnotesize
    \begin{tabular}{p{50em}}
    \toprule
   \textbf{Instruction} \\
I like to host guests at my home from time to time, and I am gathering recipes of different dishes and drinks to keep things interesting. I am interested in trying some Somali dishes. Can you give me a recipe for Canjeero?
\\
\midrule
\textbf{Output} \\
Here's a traditional and delicious Somali recipe for Canjeero (also known as Canjelo or Canjif) - a popular flatbread that's often enjoyed for breakfast or as a snack. This dish is versatile and can be customized with various toppings to suit your preferences. Here's how to make it:\newline \newline Ingredients:\newline 1. 2 cups all-purpose or bread flour\newline 2. 1/2 cup fine semolina (sooji or rava)\newline 3. 1 tsp salt\newline 4. 1 1/2 cups water (or as needed)\newline 5. 2 tbsp vegetable oil (or ghee/clarified butter)\newline 6. Toppings of your choice:\newline    \hspace*{1em} a. Yogurt\newline   \hspace*{1em} b. Fresh or dried fruits (bananas, mangoes, dates, raisins)\newline  \hspace*{1em}  c. Honey or sugar\newline  \hspace*{1em}  d. Cardamom powder\newline  \hspace*{1em}  e. Peanuts or other nuts\newline   \hspace*{1em} f. Qaat (a popular Somali leaf) - optional and not commonly found outside of Somalia\newline \newline Instructions:\newline \newline 1. In a large mixing bowl, combine the all-purpose or bread flour, semolina, and salt. Mix well to combine.\newline \newline 2. Gradually pour in the water while stirring with a wooden or plastic spoon. Be cautious while adding water, as the dough can become too sticky if excess water is added. Add water until the dough comes together as a smooth ball.\newline \newline 3. Knead the dough for 3-4 minutes until it becomes soft and elastic. Cover the bowl with a damp cloth or plastic wrap and let the dough rest for 10-15 minutes. This allows the gluten to develop, resulting in a fluffier bread.\newline \newline 4. Heat a non-stick or cast-iron skillet on medium-high heat. Add the vegetable oil (or ghee/clarified butter) and let it heat up.\newline \newline 5. Divide the dough into equal-sized balls (usually about 6-8 pieces). Flatten each ball into a disc using your palms. Make sure the disc is not too thin, as this will result in a crispier bread.\newline \newline 6. Once the skillet is hot, carefully place the flattened discs into the skillet. Do not overcrowd the skillet; cook the bread in batches if necessary. Cook the bread for 1-2 minutes on each side, or until golden brown spots appear. Press down lightly on the bread with a spatula to fluff it up.\newline \newline 7. Remove the Canjeero from the skillet and place it on a plate. Repeat the process with the remaining dough balls.\newline \newline 8. Serve the hot Canjeero with your choice of toppings. Here are some popular combinations:\newline    \hspace*{1em} a. Yogurt and honey or sugar\newline   \hspace*{1em}  b. Fresh fruit and cardamom powder\newline    \hspace*{1em} c. Qaat and peanuts\newline   \hspace*{1em} d. Peanut butter and banana\newline   \hspace*{1em} e. Honey and sliced almonds\newline  \hspace*{1em}  f. Cinnamon and raisins\newline \newline Note: Canjeero is a versatile dish that can be enjoyed any time of day. It's often served with tea or coffee as a snack or as part of a larger meal. You can also experiment with different types of flour, such as whole-wheat or spelt, to create unique flavors.\newline \newline Enjoy your homemade Canjeero! This dish is a staple in Somali cuisine and is a great way to explore the rich flavors and traditions of this vibrant culture. If you have any questions or need further assistance, please don't hesitate to ask. Happy cooking!
\\
    \bottomrule
    \end{tabular}%
\caption{Output example with the posterior probability  objective (Eq.~\ref{eq:obj-prob}).}
  \label{tab:output-probability}
\end{table*}%

\begin{table*}[ht]
  \centering
  \footnotesize
    \begin{tabular}{p{50em}}
    \toprule
   \textbf{Instruction} \\
I like to host guests at my home from time to time, and I am gathering recipes of different dishes and drinks to keep things interesting. I am interested in trying some Somali dishes. Can you give me a recipe for Canjeero?
\\
\midrule
\textbf{Output} \\
Certainly! Here's a simple recipe for Canjeero (Somali flatbread) that serves 4-6 people:\newline \newline Ingredients:\newline - 2 cups (240 grams) of Somali flour (also known as ``soor'' or ``sorghum flour'')\newline - 2 cups (480 ml) of water\newline - 1/2 teaspoon of salt\newline \newline Instructions:\newline \newline 1. In a large mixing bowl, combine the somali flour and salt.\newline \newline 2. Gradually add the water to the flour mixture while stirring with a wooden spoon or spatula until the mixture is smooth, free of lumps, and has a thick consistency.\newline \newline 3. Cover the bowl with a clean kitchen towel or plastic wrap and let it rest at room temperature for at least 1 hour (or up to 24 hours) to allow the flour to hydrate and the gluten to develop.\newline \newline 4. When ready to make the canjeero, preheat a non-stick pan or griddle over medium heat.\newline \newline 5. Using a measuring cup or ladle, pour about 1/3 to 1/2 cup of batter into the center of the pan.\newline \newline 6. Immediately, use the back of the ladle or a spatula to gently spread the batter into a thin, round shape (about 6-8 inches in diameter).\newline \newline 7. Let the canjeero cook for 1-2 minutes or until the bottom is golden brown and the top is beginning to set.\newline \newline 8. Flip the canjeero over and let it cook for another 30 seconds to 1 minute on the other side, or until golden brown.\newline \newline 9. Repeat with the remaining batter, using more oil or butter to lubricate the pan as needed.\newline \newline 10. Serve the canjeero warm with a side of ghee (clarified butter) or other Somali stews or sauces like ``Wat'' or ``Xalwo.''\newline \newline Enjoy your traditional Somali Canjeero!
\\
    \bottomrule
    \end{tabular}%
\caption{Output example with the likelihood function objective (Eq.~\ref{eq:obj-likelihood}).}
  \label{tab:output-likelihood}
\end{table*}%

\end{document}